\documentclass[10pt,letterpaper]{article}
\usepackage{graphicx} 
\usepackage{cogsci}
\usepackage{multirow} 
\usepackage[table]{xcolor} 
\definecolor{lightgray}{gray}{0.9} 
\usepackage{booktabs}  
\usepackage{makecell}  
\cogscifinalcopy 

\usepackage[
  style=apa,
  natbib=true,
  annotation=false,
]{biblatex}
\usepackage{float} 

\title{Automatic Cognitive Task Generation for In-Situ Evaluation of Embodied Agents}

\author[1,2]{\mbox{Xinyi He}}
\author[2]{\mbox{Ying Yang}}
\author[2]{\mbox{Chuanjian Fu}}
\author[1]{\mbox{Sihan Guo}}
\author[1,2]{\mbox{Songchun Zhu}}
\author[2]{\mbox{Lifeng Fan}}
\author[2]{\mbox{Zhenliang Zhang}}
\author[1,2]{\mbox{Yujia Peng}}

\affil[1]{Peking University}
\affil[2]{State Key Laboratory of General Artificial Intelligence, BIGAI, Beijing, China}

\begin{document}

\maketitle

\begin{abstract}As general intelligent agents are poised for widespread deployment in diverse households, evaluation tailored to each unique unseen 3D environment has become a critical prerequisite. However, existing benchmarks suffer from severe data contamination and a lack of scene specificity, inadequate for assessing agent capabilities in unseen settings. To address this, we propose a dynamic in-situ task generation method for unseen environments inspired by human cognition. We define tasks through a structured graph representation and construct a \textbf{t}wo-stage interaction-evolution task generation system for \textbf{e}mbodied \textbf{a}gents (\textbf{TEA}). In the interaction stage, the agent actively interacts with the environment, creating a loop between task execution and generation that allows for continuous task generation. In the evolution stage, task graph modeling allows us to recombine and reuse existing tasks to generate new ones without external data. Experiments across 10 unseen scenes demonstrate that TEA automatically generated 87,876 tasks in two cycles, which human verification confirmed to be physically reasonable and encompassing essential daily cognitive capabilities. Benchmarking SOTA models against humans on our in-situ tasks reveals that models, despite excelling on public benchmarks, perform surprisingly poorly on basic perception tasks, severely lack 3D interaction awareness and show high sensitivity to task types in reasoning. These sobering findings highlight the necessity of in-situ evaluation before deploying agents into real-world human environments.

\end{abstract}

\begin{figure*}[t]
    \centering
    \includegraphics[width=0.9\textwidth]{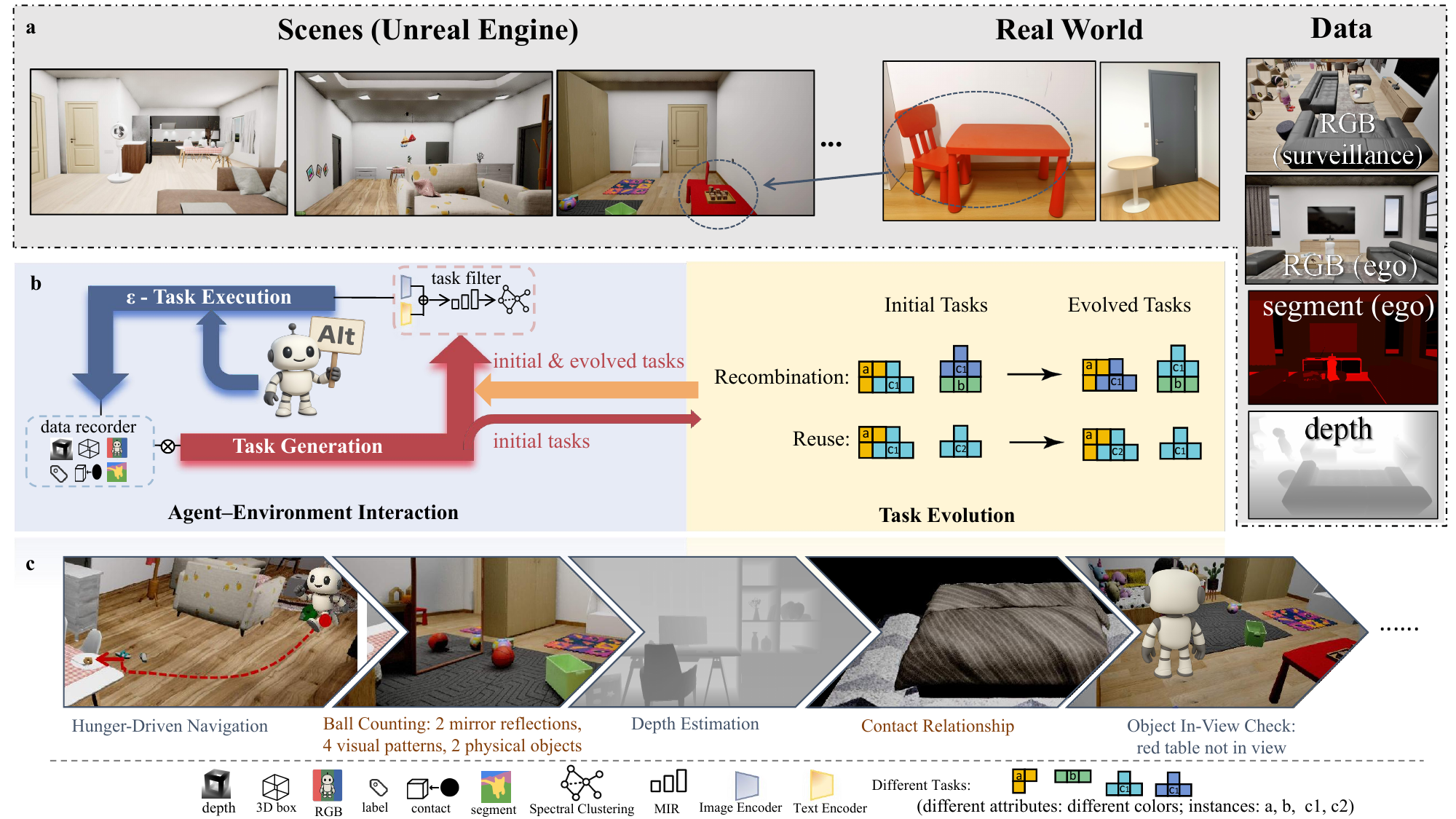}
    \caption{
Overview of two-stage dynamic task generation system and task examples. (a) Samples of UE and real-world scenes, and data returned by the UE simulator.
(b) Agent–Environment Interaction (left): 
The agent executes tasks, collects data, and generates new tasks based on recorded data. 
A task filter selects a subset of tasks for the next iteration, while an $\epsilon$-randomwalk ensures diversity. 
 Task Evolution (right): The task recombination and reuse strategy leverages existing tasks to form new ones. 
(c) Example tasks such as navigation and object in-view check (e.g., the red table is not visible from the agent’s view).
}
    \label{fig:TEA}
\end{figure*}

\section{Introduction}
\label{sec:intro}

Recent advances in large vision-language models (VLMs) have revealed remarkable potential for building embodied agents that can explore and understand the 3D world (\cite{gpt,llama,gemini}). As agents move towards real-world deployment (\cite{gao2019vrkitchen,ge2024commonsense}), the ability to automatically generate adaptive and large-scale tasks in 3D environments becomes crucial for in-situ evaluation. 

However, existing task generation techniques face limitations when conducting in-situ evaluations tailored to each
unique unseen 3D environment:
\textbf{(1) Domain Gap:} Most existing approaches, which have shown impressive performance, are designed for highly structured domains such as code-based (\cite{zhao2025absolutezero}), GUI-based (\cite{omnibench}), or game tasks (\cite{zheng2025mcu, tang2024mars}). 
However, these environments lack the complexity and physical realism required to represent embodied worlds, making direct adaptation to 3D environments infeasible.
\textbf{(2) Failure in unseen environments:} Current embodied task generation methods typically depend on initialized task instances and generate new ones based on existing tasks by injecting external objects, altering room layouts or through data augmentation (\cite{chen2024mega,deitke2022️ProcTHOR,wang2023robogen,zhang2024taskme}). This strategy is ineffective in new environments where no initial task instances are available. 
For evaluation, relying on existing data is problematic due to severe data contamination. Recent research indicates that the overlap between training sets and public benchmarks (specifically image-text contamination) reaches a staggering 33.1\% (\cite{overlap}). Methods that generate tasks by perturbing these existing datasets fail to escape this contamination, leading to inflated results that obscure the model's actual performance. 
\textbf{(3) Violation of in-situ prerequisites: }Most existing methods rely heavily on introducing massive external assets, such as injecting external objects or importing new room maps (\cite{deitke2022️ProcTHOR,wang2023robogen}). Such dependency alters the target scenes, thereby violating in-situ prerequisites. We argue that tasks themselves contain rich structural information that can be recombined and reused to create new in-situ tasks, even without additional resources.

To enable in-situ task generation in unseen embodied environments, we formulate a \textbf{t}wo-stage interaction-evolution task generation system for \textbf{e}mbodied \textbf{a}gents (\textbf{TEA}, Fig~\ref{fig:TEA}). Our system is implemented within Unreal Engine (UE), a robust platform supporting high-fidelity real-world scene scanning and accurate physics-based simulation (\cite{engine2018unreal,sun2025tongsim}). Tasks are defined as graph-structured representations inspired by human cognition, which model the varying cognitive entities, relationships, and depth required by different tasks.
The process consists of two stages, agent–environment interaction and task evolution. Prior work mainly focuses on environment variation.
For agent–environment interaction, we introduce an \textbf{agent-in-loop task generation method} that enables task generation even without initial tasks, which forms a closed loop between task generation and execution, enabling adaptive and self-improving task creation.
For task evolution, we introduce a \textbf{task recombination and reuse strategy} based on the graph structure of tasks, which regenerates tasks from existing tasks, taking good use of existing data instead of relying heavily on external environment variations.
To evaluate the quality of generated tasks, we introduce metrics, including the Maximum Independent Ratio (MIR) and spatial statistics, to quantify their variation and distribution in 3D environments.

Across 10 new scenes, our system automatically generated 87876 tasks in two iterations without any initial tasks. Our framework consistently improves task diversity (p $<$ 0.001 in t-tests).
We evaluate state-of-the-art (sota) models on our generated benchmarks (Fig~\ref{fig:task-space}). The results, when compared to human performance as shown in Table~\ref{tab:model_comparison}, reveal that model, despite excelling on public benchmarks, perform surprisingly poorly on basic perception tasks, severely lack 3D interaction awareness and show high sensitivity to task types in reasoning. These sobering findings highlight the necessity of environment-specific evaluation before deploying agents into real-world human environments. There is often a tendency to attribute human-like generalization to models, where high performance on a test set is interpreted as mastery of the underlying skill. This intuitive expectation may lead users to overestimate model's performance on similar tasks. Our results challenge this view, suggesting that relying on such assumptions could introduce risks in real-world deployment. 



\section{Methods}
In this section, we define tasks based on graph structures, formulate a two-stage framework for task generation, and introduce agent-in-loop task generation method, and graph-based task recombination and reuse methods as shown in Fig~\ref{fig:TEA}. 

\subsection{Task definition}
We define tasks in a graph-structured manner, modeling them using vertices (units to be processed, such as objects, scenes and agents), edges (relationship between vertices, such as spatial relationship and ownership) and attributes (information of vertices and edges, such as labels). This structure is cognitive-inspired, mapping vertices, edges, and attributes to cognitive entities, relationships, and cognitive depth, respectively. Prior work typically proposed tasks independently in natural language. However, semantic ambiguity often causes different tasks to share the same name, e.g., object recognition may refer either to a CNN(\cite{he2016deep}) giving probabilities over predefined categories or to a GPT model(\cite{gpt}) generating semantic labels. By formalizing tasks as graphs, we unlock the ability to decompose and reorganize task components. This allows, for example, seamlessly evolving an object detection task (output: bounding box and label) into a classification task (label only) without external data.

\begin{figure*}[t]
    \centering
    \includegraphics[width=0.9\textwidth]{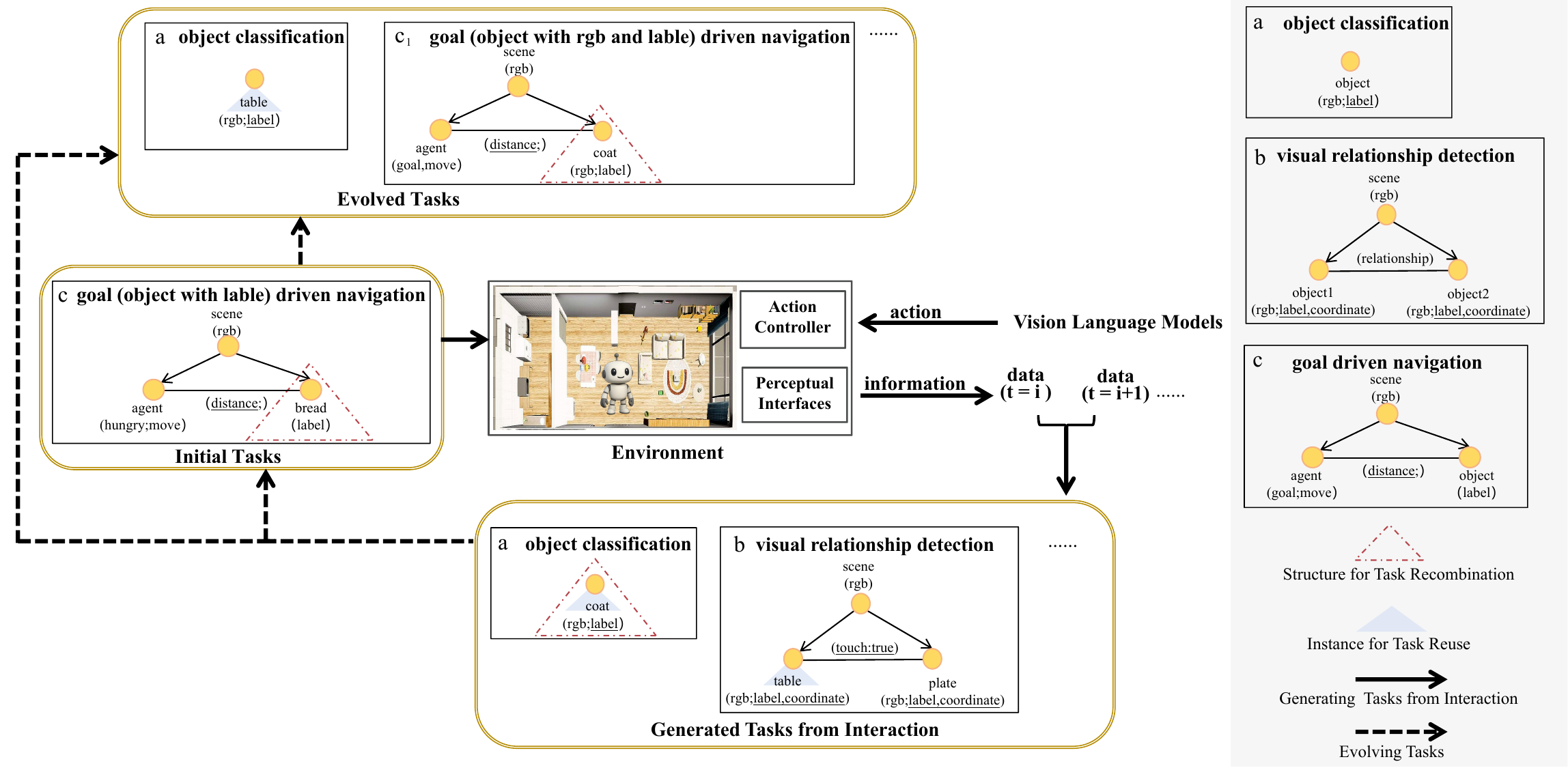}
    \caption{
        Task evolution visualization: solid arrows represent generating tasks through interaction during the execution of initial tasks, while dashed arrows represent evolving tasks based on existing tasks. For task evolution, blue triangular structures depict the reuse structure between tasks while the red dashed triangular illustrates task recombination which exchange nodes with different attributes (underlined words represent the added elements in the final state compared with the initial state).
    }
    \label{fig:task-evolve}
\end{figure*}

We organize tasks into a graph structure using \textbf{v}ertices(V), \textbf{e}dges (E) and \textbf{a}ttributes(A):
\begin{align}
\text{Task} &= ( \text{Initial-State}, \text{Final-State} ) \\
\text{Initial-State} &= ( V, E, A(V, E) ) \\
\text{Final-State} &= ( V, E, A(V, E) )
\end{align}
For example, the initial state of an object classification task is (\{object\}, $\varnothing$, \{color picture (object)\}) and final-state is (\{object\}, $\varnothing$, \{color picture (object), \underline{label (object)}\}). The underlined part represents the added elements compared with the initial state. For simplicity, we use the underlined final-state to denote the entire task. This is an abstract description of a task as it does not refer to any specific executable task. When the graph is mapped to a concrete environment, we obtain a specific task instance. For example, if we map the object to an apple, then (\{apple A\}, $\varnothing$, \{color picture (apple A), \underline{label (apple A)}\}) represents a concrete task \textbf{instance}. As shown in Fig~\ref{fig:task-evolve}, the left part shows task \textbf{instances} and the right shows abstract tasks. 

\begin{figure*}[h]
    \centering
    \includegraphics[width=0.8\textwidth, trim=0.2cm 0.2cm 2.5cm 0.2cm, clip]{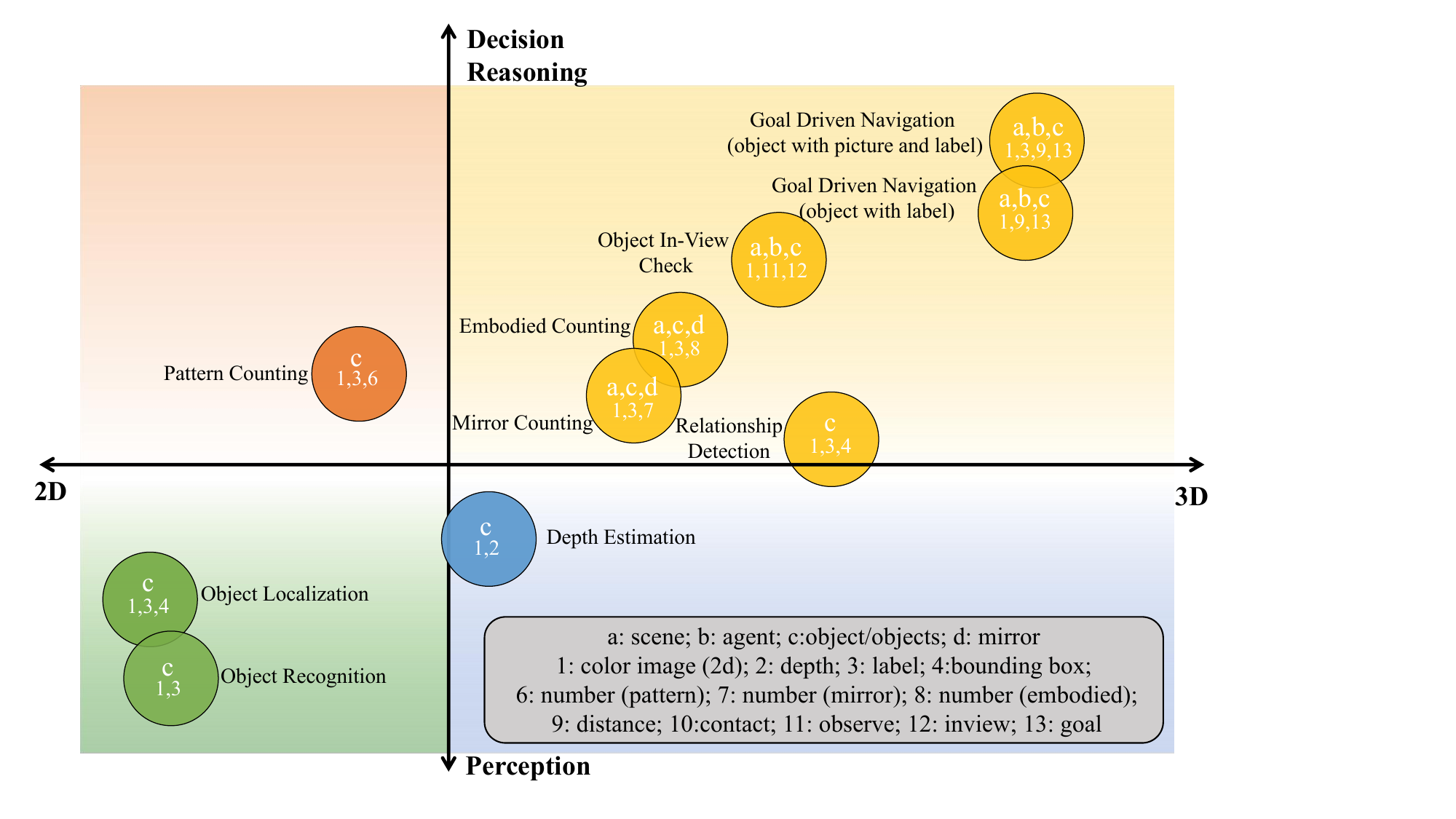}
    \caption{\textbf{A Taxonomy of Tasks.} Tasks are organized by  dimension (2D image vs. 3D physical space) and cognitive load (from perception to reasoning and decision-making).}
    \label{fig:task-space}
\end{figure*}


\subsection{A Two-Stage Task Generation Method}
We formulate a two-stage framework for task generation, where the system takes the environment (env) and agent as inputs, and outputs task instance set $\boldsymbol{\tau}$:
\begin{equation}
    \boldsymbol{\tau} = \mathrm{Evo}\big( T(\mathrm{env}, \mathrm{agent});\ \beta \big).
\end{equation}
The process consists:
(1) Agent–Environment Interaction: $T(\mathrm{env}, \mathrm{agent})$ describes how the agent interacts with the given environment to generate new tasks. In this stage, we introduce an \textbf{agent-in-loop task generation method}, which forms a closed loop between task generation and execution, enabling adaptive and continuous task creation even without initial task instances. The agent initiates a random exploration to form a self-driven loop between task execution and task generation in the absence of task instances.
(2) Task Evolution: $\mathrm{Evo}(\tau;\beta)$ regenerates tasks based on existing tasks $\tau$ according to evolutionary rules $\beta$.
In this stage, we introduce a \textbf{task recombination and reuse strategy} based on the graph structure of tasks, effectively leveraging existing data to synthesize new tasks.



\begin{figure*}[t]
    \centering
    \includegraphics[width=\textwidth, trim=0cm 10.7cm 10cm 0cm, clip]{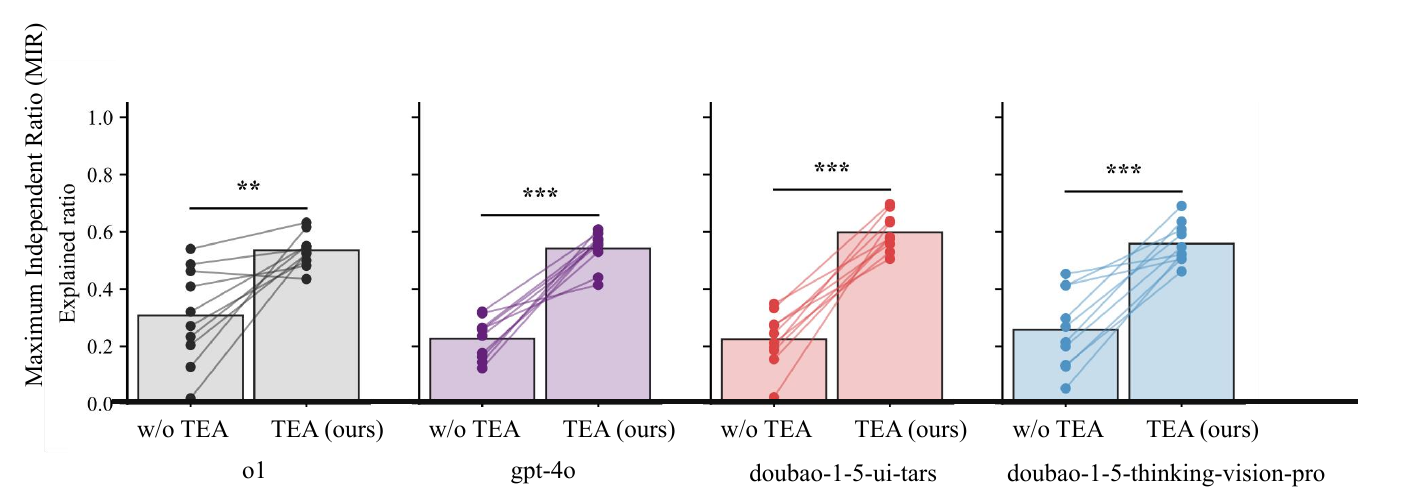}
    \caption{
Comparison of MIR across four VLMs and 10 different scenes.
Each point represents one scene and bars show the average MIR over ten scenes.
Our methods consistently improve task diversity (t-tests: $^{**}$\,$p<0.01$, $^{***}$\,$p<0.001$).
    }
    \label{fig:mir-t}
\end{figure*}

\subsubsection{Agent-in-Loop Task Generation Method}
 
We propose an agent-in-loop task generation method, where the agent can generate new tasks during task execution, enabling continuous task generation.
$\boldsymbol{\tau}$ denotes the task set, where each $t \in \boldsymbol{\tau}$ represents a single task instance.  
$\mathbf{D}$ denotes the set of data samples (i.e., images, positions and labels) returned from UE. $\boldsymbol{\tau} $ is generated by data $\mathbf{D}$, using the function {G} which contains various task functions $g_i$ (e.g., generating a classification task instance from an object image and its label). The receiving function $R$ collects data $\mathbf{D}$ through continuous agent-environment interaction, receiving real-time scene information after each agent action. With a small probability $\varepsilon$, the agent executes random walk to encourage exploration (
$t^{\mathsf{inter}}$ denotes the task requiring interaction with the environment and $t^{\mathrm{rw}}$ denotes a random walk process controlled by random numbers):
\begin{align}
\boldsymbol{\tau}
  &= {G}(\mathbf{D})  
  = \underset{g_i \in \mathbf{G}}{\bigcup} \, {g_i}(\underset{d \in \mathbf{D}}{\bigcup}d) 
   = \{t_{1},\, t_{2},\, \ldots,t_{n}\}
  \label{eq:tau-generation} \\[8pt]
\mathbf{D}^{\prime}
  &= \begin{cases}
    \underset{t \in \boldsymbol{\tau}^{\mathsf{inter}}}{\bigcup}
    {R}(t,\, \text{agent}),
    & \text{with probability } 1 - \varepsilon, \\[6pt]
    {R}(t^{\mathrm{rw}},\, \text{agent}),
    & \text{with probability } \varepsilon,
  \end{cases}
  \label{eq:data-update}
\end{align}
\noindent
The updated task set is then given by
$\boldsymbol{\tau}^{\prime}
  =  {G}(\mathbf{D}^{\prime})$. This cyclical procedure enables adaptive task generation and continual data accumulation (in special cases where $\tau$ = $\varnothing$, the method can still operate by setting $\varepsilon = 1$ in Equation~\ref{eq:data-update} to initiate the entire loop):
\begin{align}
\mathbf{D}
\;\xrightarrow{\;G\;}\;
\boldsymbol{\tau}
\;\xrightarrow{\;R\;}\;
\mathbf{D}^{\prime}
\;\xrightarrow{\;G\;}\;
\boldsymbol{\tau^{\prime}}
\;\xrightarrow{\;\;}\;
\cdots
\label{eq:loop}
\end{align}

\noindent\textbf{Task Filter.} Executing all tasks from the previous loop would cause an exponential increase in the number of tasks for subsequent loops. To prevent exponential explosion and ensure diversity with minimal redundancy, we introduce a task filter to select representative tasks.
Given a candidate task set $\boldsymbol{\tau} = \{t_1, t_2, \ldots, t_n\}$, we compute a pairwise similarity matrix $S \in \mathbb{R}^{n \times n}$ 
using multi-modal embeddings extracted from vision–language models:
\begin{align}
S_{ij} &= \mathrm{sim}(t_i, t_j)
= \frac{1}{n}\sum_{m=1}^{n}\cos\bigl(f_m(t_i), f_m(t_j)\bigr),
\label{eq:sim-matrix}
\end{align}
where $f_m(\cdot)$ denotes the encoder for the $m$-th modality (i.e., image, command prompt, and label).
We apply spectral clustering to \( S \) to partition tasks into \( k \) clusters.  
The representative task set is obtained as:
$\boldsymbol{\tau}^{*} = \{ t^{*}_1, t^{*}_2, \ldots, t^{*}_K \}$, where \( t^{*}_i \) denotes the representative task of the \( i \)-th cluster.

\begin{table*}[ht]
\centering
\caption{MIR comparison for VLM-generated task subsets. We compare the \textbf{ablation version without our method} ($\boldsymbol{\tau}_{\text{without}}$, \textcolor{gray}{gray}) against our proposed full method: the \textbf{first-round tasks} ($\boldsymbol{\tau}$) and the \textbf{second-round tasks} ($\boldsymbol{\tau}'$). Each cell displays the MIR value followed by (MIS count / Total count).}
\small
\begin{tabular}{
c|c@{\hspace{1pt}}c|c@{\hspace{1pt}}c|
c@{\hspace{1pt}}c|c@{\hspace{1pt}}c|c@{\hspace{1pt}}c|c@{\hspace{1pt}}c|c@{\hspace{1pt}}c
}
\toprule
\multirow{2}{*}{\textbf{Scene}} &
\multicolumn{2}{c|}{\textbf{\multirow{2}{*}{\textcolor{gray}{$\boldsymbol{\tau_{\text{without}}}$}}}} &
\multicolumn{2}{c|}{\textbf{\multirow{2}{*}{$\boldsymbol{\tau}$ (ours)}}} &
\multicolumn{10}{c}{\textbf{$\boldsymbol{\tau'}$ (ours)}} \\
\cmidrule(lr){6-15}
& & & & &
\multicolumn{2}{c|}{$\boldsymbol{\tau}_1^{\prime}$} &
\multicolumn{2}{c|}{$\boldsymbol{\tau}_2^{\prime}$} &
\multicolumn{2}{c|}{$\boldsymbol{\tau}_3^{\prime}$} &
\multicolumn{2}{c|}{$\boldsymbol{\tau}_4^{\prime}$} &
\multicolumn{2}{c}{$\boldsymbol{\tau}_5^{\prime}$} \\
\midrule
0 & \textcolor{gray}{0.13}&\textcolor{gray}{(11/86)} & 0.62 & (53/86)&      1.00 & (6/6) & 0.73 & (38/52) & 0.80 & (4/5) & 0.73 & (43/59) & 0.67 & (4/6) \\
···\\
9 & \textcolor{gray}{0.02} &  \textcolor{gray}{(2/108)}& 0.53 & (57/108)  & 0.32 & (17/53) & 1.00 & (6/6) & 0.66 & (21/32) & 0.67 & (18/27) & 0.60 & (18/30) \\

\midrule
\midrule
mean±std &   \textcolor{gray}{0.307}& \textcolor{gray}{± 0.159}& \ 0.536  & ± 0.055  & \multicolumn{10}{c}{0.676 ± 0.156}\\
\bottomrule
\end{tabular}
\label{tab:mir_compare}
\end{table*}

\subsubsection{Structure-based Task Recombination and Reuse}
We propose a task recombination and reuse strategy that fully leverages existing tasks inspired by human creativity process (\cite{peng2025probing}), allowing tasks to evolve into new ones via the flexible disassembly and reassembly of their graph structures. As shown in Fig~\ref{fig:task-evolve}, where we take three types of tasks ($t^{\mathrm{a}}$, $t^{\mathrm{b}}$, $t^{\mathrm{c}}$) as examples, we distinguish between two generation pathways: solid arrows represent tasks generated via real-time interaction (during the execution of initial tasks), while dashed arrows represent tasks evolved from existing structures.

\noindent\textbf{Task Reuse.} 
The blue triangular regions in Fig.~\ref{fig:task-evolve} illustrate the reuse structures.
We introduce the notation $t^{\mathrm{a}} \preceq t^{\mathrm{b}}$ to denote that the graph structure of task $t^{\mathrm{a}}$ is a substructure of task
$t^{\mathrm{b}}$. Under the condition $t^{\mathrm{a}} \preceq t^{\mathrm{b}}$,
task $t^{\mathrm{a}}$ can directly inherit valid instances associated with the shared substructure from $t^{\mathrm{b}}$. For example, the classification task can
reuse the \emph{table} instance already identified within a more complex visual relationship
detection task instance.

\noindent\textbf{Task Recombination.} 
The red dashed triangular structures in Fig~\ref{fig:task-evolve} illustrate task recombination. This mechanism generates new task templates by exchanging vertices that share the same semantic type (e.g., both representing objects) but differ in attributes. Swapping their vertex structure leads to new task structures. For instance, a label-based object searching task can be recombined into a vision-based searching task by replacing the label-conditioned object node with an image-conditioned object node, thereby generating a new task template.

\section{Experiments}
We conduct two sets of experiments to evaluate generated tasks, and test models and humans on the generated tasks.

\noindent\textbf{Experimental Setup.}
Due to the large number of generated tasks, we execute two loops (first loop $\varepsilon = 1$ and second loop $\varepsilon = 0$) and analyze $\boldsymbol{\tau}$ and $\boldsymbol{\tau^{\prime}}$. To ensure diversity, the loop is executed from ten different unseen scenes, with k = 5 in the task filter. 
The agent is controlled by specifying the walking direction and distance, during which we collect RGB, instance segmentation, and depth images from both ego and surveillance views as shown in Fig~\ref{fig:TEA}a. In addition, we record labels, positions and 3D bounding boxes of objects. Each record is timestamped. 
Based on the collected data, task generation employs hybrid strategies: rule-based generation is used for deterministic tasks (e.g., counting, localization) without external models, while VLMs facilitate the generation of semantic tasks (e.g., navigation with language commands)
\subsection{Experiment 1: Generated Tasks}
We present metrics to evaluate the two stages and validate the quality of tasks.

\noindent\textbf{MIR: Maximum Independent Ratio.}
LLMs tend to produce long and verbose outputs, leading to high task redundancy and low diversity, where tasks are densely clustered in the same space.  
To quantify this redundancy, we introduce the Maximum Independent Ratio (MIR) to evaluate the proportion of non-redundant tasks in a given task set.  
Formally, given a redundancy threshold $\alpha \in (0, 1)$, we define the Maximum Independent Subset (MIS) as the largest subset 
$\boldsymbol{\tau}' \subseteq \boldsymbol{\tau}$ that maximizes the number of retained tasks under the independence constraint:
\begin{align}
\mathrm{MIS}(\alpha, \boldsymbol{\tau})
&= \underset{\boldsymbol{\tau}' \subseteq \boldsymbol{\tau}}{\arg\max}
\Bigl\{\,|\boldsymbol{\tau}'|
\;\Big|\;
\forall\, t_i \neq t_j \in \boldsymbol{\tau}',\;
S_{ij} \le \alpha
\Bigr\}.
\label{eq:mis}
\end{align}
The Maximum Independent Ratio (MIR) is then defined as the ratio between the size of this subset and the total number:
\begin{align}
\mathrm{MIR}(\alpha, \boldsymbol{\tau})
&= \frac{|\mathrm{MIS}(\alpha, \boldsymbol{\tau})|}{|\boldsymbol{\tau}|}.
\label{eq:mir}
\end{align}
A higher MIR indicates greater task diversity and lower redundancy. We set $\alpha = 0.8$ in this paper.

\begin{table}[h]
\centering
\caption{
Spatial statistics of embodied task instances across $\boldsymbol{\tau}$ and $\boldsymbol{\tau}'$, including 
the enclosing-space volume ($V_{\text{all}}$ in m$^3$), 
per-axis range/standard deviation ($\Delta$/$\sigma$ in m), 
mean$\pm$std of instance volumes ($\bar{V}_{\text{inst}}$ in m$^3$), 
and object count ($N_{\text{obj}}$).
}

\small
\begin{tabular}{c|c|cc}
\toprule
\textbf{metrics} & \textbf{$\boldsymbol{\tau}$} & \textbf{$\boldsymbol{\tau_1'}$} & \textbf{$\boldsymbol{\tau_2'} $}  ···\\
\midrule
$V_{\text{all}}$  & 744.84 & 744.84 & 744.84 \\
\rowcolor{gray!10}
$\Delta_x / \sigma_x$  & 9.37 / 1.84 & 9.37 / 2.00 & 9.37 / 1.91 \\
\rowcolor{gray!10}
$\Delta_y / \sigma_y$  & 20.90 / 1.94 & 20.90 / 1.94 & 20.90 / 1.92 \\
\rowcolor{gray!10}
$\Delta_z / \sigma_z$  & 3.80 / 0.92 & 3.80 / 1.01 & 3.80 / 0.97 \\
$\bar{V}_{\text{inst}} \pm \sigma$  & 18.51 $\pm$ 58.26 & 43.60 $\pm$ 93.30 & 24.43 $\pm$ 69.04 \\
\rowcolor{gray!10}
$N_{\text{obj}}$ & 38 & 19 & 34 \\

\bottomrule
\end{tabular}

\label{tab:spatial_stats_tauprime2}
\end{table}

\begin{table*}[htbp]
\centering
\caption{Comparison of accuracy and mIoU across models (\cite{gpt, llama, gemini}).
The highest result per task is \textbf{bolded}, the second highest is \underline{underlined}, 
and the lowest is \textcolor{gray}{grayed out}. $\Delta$ denotes the gap between the highest and lowest scores.}
\label{tab:model_comparison}
\resizebox{\textwidth}{!}{
\begin{tabular}{lccccccccc|c}
\toprule
\textbf{Task Type} & \textbf{GPT-5} & \textbf{GPT-4o} & \textbf{GPT-4.1} & \textbf{Gemini-2.5-Pro} & \textbf{Gemini-2.5-Flash} & \textbf{Claude-Opus-4} & \textbf{llama-4-Scout} & \textbf{o3-2025} & \textbf{Human} & $\boldsymbol{\Delta}$ \\
\midrule
Embodied Counting      & \textcolor{gray}{0.6571} & 0.7810 & 0.8000 & 0.7143 & \underline{0.8190} & 0.7143 & 0.7619 & 0.8095 & \textbf{0.8476} & 0.1905 \\
Mirror Counting        & 0.8800 & 0.9200 & 0.8800 & 0.8800 & 0.8700 & \underline{0.9700} & \underline{0.9700} & \textbf{0.9800} & \textcolor{gray}{0.8300} & 0.1500 \\
Object In-View Check   & 0.4872 & 0.4274 & \textcolor{gray}{0.1880} & \underline{0.4957} & 0.4530 & 0.4274 & 0.4444 & 0.4359 & \textbf{0.6239} & 0.4359 \\
Object classification     & 0.4056 & 0.0769 & 0.3566 & 0.3357 & 0.3217 & \textcolor{gray}{0.0490} & 0.3007 & \underline{0.4196} & \textbf{0.7448} & 0.6958 \\
Object Localization & 0.3174 & \textcolor{gray}{0.0468} & 0.2396 & 0.1587 & 0.1242 & \underline{0.4367} & 0.0775 & 0.3703 & \textbf{0.4913} & 0.4445 \\
Pattern Counting       & 0.7010 & \textcolor{gray}{0.6701} & 0.7216 & 0.6804 & \underline{0.7938} & 0.7423 & 0.7010 & 0.7835 & \textbf{0.8041} & 0.1340 \\
Relationship Detection  & \textcolor{gray}{0.8050} & 0.8300 & 0.8500 & 0.8300 & 0.8600 & 0.8700 & 0.8100 & \textbf{0.9400} & \underline{0.8900} & 0.1350 \\
\bottomrule
\end{tabular}
}
\end{table*}

\begin{table}[htbp]
\centering
\small 
\caption{Performance comparison on navigation tasks across GPT-4o and o1. Parentheses indicate the range of each metric ($\uparrow$: higher is better).}
\begin{tabular}{l|ccccc}
\toprule
model & \makecell{Nav.\\Gain  \\ \scriptsize{[-1, 1]$\uparrow$}} & \makecell{Success\\Rate  \\ \scriptsize{[0, 1]$\uparrow$}} & \makecell{Step\\ Number  \\ \scriptsize{[0, 10]$\downarrow$}} & \makecell{Target \\Negl. Rate  \\ \scriptsize{[0, 1]$\downarrow$}} & \makecell{Lack of\\ 3D Aw.
  \\ \scriptsize{[0, 1]$\downarrow$}} \\
\midrule
GPT-4o & \textbf{0.48} & \textbf{0.62} & \textbf{4.04} & \textbf{0.08} & 0.14 \\
o1    & 0.12 & 0.38 & 9.80 & 0.72 & \textbf{0.00} \\
\bottomrule
\end{tabular}
\label{tab:performance_comparison_navi}
\end{table}

\noindent\textbf{Interaction Stage.}  
Since VLMs fail on rule-driven tasks (e.g., counting and segmentation), we only perform our MIR analysis on tasks reliant on VLM-generated instructions. To ensure a fair comparison, we provide identical initial images and prompts for both task generation without our method (w/o TEA) and our TEA-based approach.
This ensures that the denominator of the MIR remains consistent and that the total number of generated tasks is controlled for a fair comparison.

As shown in Fig~\ref{fig:mir-t}, our method consistently achieves higher MIR scores across 4 VLMs in 10 scenes.  
Each point represents one scene and lines connecting paired points indicate the within-scene comparison. 
We further evaluate the second loop under the same setting.
Table \ref{tab:mir_compare} compares MIR for the first-round tasks $\tau$, the first-round tasks without our method  $\boldsymbol{\tau_{\text{without}}}$,, and the second-round tasks $\boldsymbol{\tau'}$, using GPT-4o. Both $\boldsymbol{\tau}$ and $\boldsymbol{\tau'}$ achieve higher MIR than $\boldsymbol{\tau_{\text{without}}}$, confirming the effectiveness of the TEA method.
$\boldsymbol{\tau'}$ shows slightly higher MIR but larger variance due to differences in navigation task complexity and agent performance, as some easy tasks end in a few steps. 

\noindent\textbf{Evolution Stage.}  
To further examine the effectiveness of the proposed task evolution mechanism,
we analyze the extent to which evolved tasks are integrated into the existing task space under the same experimental setting ($\alpha = 0.8$, task set $\boldsymbol{\tau}$).
Specifically, we take the union of the initial task set and the evolved task set, and quantify the increment in the Maximum Independent Subset (MIS) compared to the initial set.
We then compute the proportion of these newly added elements relative to the evolved task set,
which is \textbf{MIR-e}:
\begin{equation}
\mathrm{MIR\text{-}e}
=
\frac{
|\text{MIS}(\alpha, \boldsymbol{\tau}_{\text{initial}} \cup \boldsymbol{\tau}_{\text{evolve}})| 
-
|\text{MIS}(\alpha, \boldsymbol{\tau}_{\text{initial}})|
}{
|\text{MIS}(\alpha,  \boldsymbol{\tau}_{\text{evolve}})|
}.
\label{eq:mir-e}
\end{equation}
Across all scenes and models, the average \textbf{MIR-e} reaches 0.75 (GPT-4o: 0.680, GPT-o1: 0.744, Doubao-tars: 0.829, and Doubao-think: 0.731), indicating that most evolved tasks are accepted when integrated into the existing task sets.

\noindent\textbf{Spatial Statistics of Embodied Tasks.}  
We introduce spatial statistics that quantify the coverage, distribution, and diversity of embodied tasks in 3D space.
Specifically, we record the 3D positions and spatial bounding boxes of all task-related objects,
from which we derive four key metrics:
(1) $V_{\text{all}}$ (m$^3$): the total enclosing volume covered by all task instances,
reflecting the global spatial coverage of the task set;
(2) $(\Delta_x, \sigma_x)$, $(\Delta_y, \sigma_y)$, $(\Delta_z, \sigma_z)$ (m): the per-axis spatial range and standard deviation,
capturing the spread and anisotropy of task distributions;
(3) $\bar{V}_{\text{inst}}$ and $\sigma_V$ (m$^3$): the mean and variance of task instance's volumes;
(4) $N_{\text{obj}}$: the number of distinct objects involved in task set $\tau$.
Table~\ref{tab:spatial_stats_tauprime2} reports the spatial statistics for the first scene due to space constraints.  
Tasks generated during random walks (\(\boldsymbol{\tau}\)) show broader spatial coverage. 
In contrast, tasks generated during execution (\(\boldsymbol{\tau'}\)) concentrate more heavily in key functional areas of the scene and exhibit more structured spatial distributions.  
This distribution shift is desirable: the system focuses task generation on semantically meaningful regions where embodied interactions typically occur, while the small \(\varepsilon\)-probability exploration ensures that the global space remains sufficiently covered.  

\noindent\textbf{TEA-Data-88k}  
The generated dataset consists of 87,876 tasks generated from 10 scenes over only two loops, encompassing diverse task types shown in Fig~\ref{fig:TEA}c and Fig~\ref{fig:task-space}: perception (object classification, localization, depth estimation), reasoning (mirror counting, embodied counting, pattern counting), spatial reasoning (relationship detection, object in-view check), and interaction (label- and picture-driven navigation).
We validate task quality through human evaluation on a random 10\% subset of the testset. Participants indicated that all the tasks are valid (due to platform's physical constraints and strict task generation rules), 90.8\%  provide meaningful assistance at home, 84.9\% are closely related to daily routines 
and 94.4\% are identified as requiring essential cognitive faculties.
\subsection{Experiment 2: TEA-Test}
To explore models' performance on generated tasks, we sample a subset of the tasks and form a testset of 848 tasks.
Table~\ref{tab:model_comparison} compares the performance of sota models and humans (7 participants to establish a baseline). 
Furthermore, we evaluate 100 navigation tasks (shown in Table~\ref{tab:performance_comparison_navi}), focusing on only 2 representative models due to high computational cost of interaction. 
Four observations can be drawn:

\noindent\textbf{1) Basic perception tasks remain challenging for models.}  
Although classification and localization appear to be basic tasks, they exhibit the largest performance gaps ($\Delta \approx$ 0.70 and 0.44, respectively), with humans far surpassing all models.  
It shows that the bottleneck lies in VLMs’ inherent weaknesses in low-level perception, rather than in the tasks’ difficulty.
The finding highlights that basic task generation remains crucial, correcting the misconception that basic perception is a solved problem for these powerful models, while research increasingly emphasizes high-level reasoning.
\noindent\textbf{2) Humans exhibit higher hallucinations than models when reading questions.}  
Surprisingly, humans achieved the lowest score in mirror counting, a task involving counting reflected objects within mirror regions.  
Participant feedback suggests that humans often ignored the ``mirror-only'' constraint and instead counted the real-world object instances, treating the task as embodied counting.  
In some cases without visible mirrors, they still reported the actual number of objects rather than zero.  
In contrast, models, although less flexible, followed the textual instruction more literally and achieved higher accuracy. Moreover, because most scenes contain no mirrors and the correct answer is often zero, models score particularly high on this task.
\noindent\textbf{3) Reasoning task performance varies widely, revealing a bias toward training distributions.}  
Models performed best on relationship detection, a conventional reasoning task, achieving the highest accuracy (0.94) and the smallest $\Delta$ (0.14).  
However, in the newly proposed object in-view check task, which requires reasoning about the agent’s egocentric view, all models scored significantly lower than humans.  
This suggests that VLMs may perform well on reasoning types frequently seen in training data but struggle to generalize to novel embodied reasoning scenarios that demand spatial understanding and 3D awareness.
\noindent\textbf{4) Severely lack 3D interaction awareness.} We evaluate navigation across multiple dimensions shown in Table \ref{tab:performance_comparison_navi}. The ability to successfully locate the target (Success Rate) within the 10-step limit remains low even for advanced VLMs, and the navigation gain (Nav. Gain) indicates that models struggle to improve their starting position relative to the target. Failure cases reveal distinct deficits in 3D interaction awareness: o1 exhibits a high \textit{Target Neglect Rate}, frequently moving away from the target even when spotted, whereas GPT-4o primarily suffers from deficits in 3D spatial awareness (Lack of 3D Aw.), 
prematurely ending the search and interaction in 3D world if the target is not initially visible, or moving towards mirror reflections instead of turning to the actual object.




\section{Discussion and Conclusion}

In this paper, we present a unified framework and platform for automated task generation for in-situ evaluation of embodied agents. We formally define tasks as graph structures, and propose a two-stage task generation methods that enables embodied, structured and in-situ task generation. 
Evaluation results reveal that models, despite excelling on public
benchmarks, perform poorly on our in-situ tasks. These counter-intuitive results caution against the reliance on public benchmarks, urging us to reassess model capabilities within their specific deployment environments. Consequently, in-situ task generation and evaluation prove important for developing embodied agents. Due to current robotic limited performance, our evaluation focuses on agents' perception, reasoning and decision within UE. 
We will release our data 
to facilitate future research in embodied intelligence and in-situ task generation.

\nocite{August2007}
\nocite{DaphneEcho2022}
\nocite{FitzgeraldGalli1985}
\nocite{Hakuole2001}
\nocite{Issa1963}
\nocite{Lobsang2023}
\nocite{MitanniNovember1972}

\printbibliography

\end{document}